\documentclass[a4paper,fleqn]{cas-sc}

\usepackage[utf8]{inputenc}
\usepackage[T1]{fontenc}
\usepackage{amsmath,amssymb,amsfonts,bbm}
\usepackage{booktabs,graphicx,makecell,tabularx,multirow,rotating,caption}
\usepackage{microtype,xcolor,placeins,url}
\usepackage[numbers,sort&compress]{natbib}
\newcommand{\best}[1]{\ensuremath{\boldsymbol{#1}}}
\newcommand{\NA}{\textit{N}}
\newcommand{\Risk}{\textit{R}}
\newcommand{\Three}{\textit{T}}
\newcommand{\Align}{\textit{A}}

\begin{document}
\let\WriteBookmarks\relax
\def\floatpagepagefraction{.8}
\def\textpagefraction{.08}
\shorttitle{CHASE: Competing Hypotheses for Ambiguity-Aware Selective Prediction}
\shortauthors{K. Jhawar et~al.}

\title[mode=title]{CHASE: Competing Hypotheses for Ambiguity-Aware Selective Prediction}

\author[1,2]{Kartik Jhawar}
\author[1]{Yuhao Geng}
\author[1,3,4]{Atul N. Parikh}
\author[1,2]{Lipo Wang}
\cormark[1]
\ead{ELPWang@ntu.edu.sg}

\affiliation[1]{organization={Institute for Digital Molecular Analytics and Science, Nanyang Technological University},
  city={Singapore}, postcode={636921}, country={Singapore}}
\affiliation[2]{organization={School of Electrical and Electronic Engineering, Nanyang Technological University},
  city={Singapore}, postcode={639798}, country={Singapore}}
\affiliation[3]{organization={School of Materials Science and Engineering, Nanyang Technological University},
  city={Singapore}, postcode={639798}, country={Singapore}}
\affiliation[4]{organization={Singapore Centre for Environmental Life Sciences Engineering, Nanyang Technological University},
  city={Singapore}, postcode={637551}, country={Singapore}}
\cortext[cor1]{Corresponding author.}

\begin{abstract}
Selective prediction under partial observability requires distinguishing ordinary prediction error from structurally insufficient evidence. We introduce CHASE, a framework that compares class-conditional temporal hypotheses and learns when to commit or abstain. We evaluate hidden-connectivity inference in a controlled giant-unilamellar-vesicle-inspired simulator. To isolate hypothesis comparison from ambiguity supervision, we evaluate direct hypothesis-margin thresholding, error-only selectors, and matched ambiguity-supervised baseline selectors. Across ten independently generated simulator datasets, the raw hypothesis margin attains selective risk comparable to Deep Ensemble at 80\% coverage (3.80\% versus 3.82\%) while improving three-way accuracy by 1.90 percentage points (pp) and abstain alignment by 4.03 pp. Under identical ambiguity supervision at 80\% coverage, the hypothesis selector improves three-way accuracy by 9.75 pp and abstain alignment by 28.3 pp over the matched ambiguity-supervised standard-ensemble selector, with nearly identical mean selective risk. On the very-high-ambiguity subset, it reduces risk by 3.46 pp and improves three-way accuracy by 19.1 pp at 80\% coverage. The overall three-way-accuracy and abstain-alignment gains remain significant after two-sided paired tests with Holm correction and retain their direction in both leave-one-regime-out evaluations. Shuffled-label, ambiguity-threshold, and margin controls support the interpretation that the hypothesis margin ranks simulator-defined structural ambiguity. In a frozen-model applicability pilot on 11 3D-validated real-GUV cases, CHASE accepts eight cases (72.7\% coverage) with no accepted physical-label errors and abstains on the sole disconnected case, which the forced classifier misclassifies as connected. Overall, CHASE improves abstention placement at a competitive selective-risk boundary.
\end{abstract}


\maketitle

\section{Introduction} \label{intro}

Real-world decisions are often difficult not because signals are globally weak, but because they are locally misleading. A system may show short periods of strong agreement, short periods of disagreement, or partial visual evidence that appears and disappears over time. Because standard classifiers are forced to make confident binary decisions from this incomplete evidence, selective prediction is crucial: a model must learn to commit only when evidence is sufficient, and abstain otherwise.

Existing approaches usually estimate uncertainty from a single predictive branch, either by thresholding confidence, measuring predictive variability, or learning a reject option jointly with classification \citep{geifman2017selective, gal2016dropout, lakshminarayanan2017deepensembles, geifman2019selectivenet, liu2019deepgamblers, corbiere2019confidence}. These are strong and necessary baselines, but each scores uncertainty from the output of a single predictive branch, which becomes a limitation when the evidence is only partially observable and supports competing explanations over time. In such settings, the difficulty is not simply low confidence. The real difficulty is that short temporal segments can look convincing in different ways, and global summary cues can become misleading. A model should therefore not only ask ``how confident am I?'', but also ``which explanation is better supported, and is that difference reliable enough to justify commitment?''.

In this paper, we study \emph{hypothesis-driven selective prediction}. Instead of assigning confidence to one discriminative branch, CHASE compares two class-conditional temporal explanations (connected and disconnected dynamics). Clear evidence produces a reliable separation between their sequence-level negative log-likelihoods, whereas partial observability tends to collapse that separation. A learned selector can then combine prediction-error ranking with ambiguity-aware ranking to decide whether the evidence supports commitment.

The novelty of CHASE lies not in any single component but in the combination of their components and in the controlled decomposition that it enables. Class-conditional generative heads, margin training, and learned selectors each exist in prior work; what is non-obvious is that using next-frame prediction accuracy as the competition criterion between hypotheses produces a margin that is directly informative about structural ambiguity without an explicit ambiguity model or ambiguity labels at the raw-margin stage. Prior multi-hypothesis methods such as \citep{rupprecht2017learning} produce multiple outputs but do not decide between them; prior selective classifiers typically score uncertainty from a single branch without explicitly distinguishing low-confidence predictions caused by ambiguous evidence from those caused by distributional noise. CHASE bridges these two lines of work by coupling a hypothesis-margin representation to a cost-aware accept/reject policy that is explicitly trained to separate prediction error from structural insufficiency. Critically, the raw hypothesis margin alone, with no learned selector and no privileged ambiguity labels, already attains nearly identical mean selective risk to the competitive baseline while improving abstention placement, which supports the conclusion that the hypothesis representation provides complementary abstention information beyond the evaluated standard representations.

We ground the framework in hidden connectivity inference from videos of giant unilamellar vesicles (GUVs). GUVs are cell-sized lipid vesicles that are often used as simplified physical models of membrane systems. Under osmotic stress, GUVs deform and can either remain tethered in 3D by a microscopic neck or separate completely. Our task is to classify this binary connectivity from 2D projections where the physical neck (bridge) can be weak, intermittent, out of plane, or invisible. This provides a natural testbed for ambiguity-aware selective prediction, because motion correlation, proximity, and bridge intensity can disagree. Our primary quantitative benchmark is evaluated on a controlled, physics-inspired GUV simulator across ten independently generated datasets, providing ground-truth ambiguity control and rigorous statistical inference. To test applicability beyond simulation, we additionally apply the frozen simulator-trained pipeline, without retraining or threshold recalibration, to real-GUV cases whose physical connectivity is established by three-dimensional reconstruction after fixed label-blind geometric canonicalization.

In summary, this paper makes four contributions:
\begin{itemize}
    \item We formulate selective prediction through explicit competition between class-conditional temporal hypotheses, yielding a direct likelihood-margin abstention signal.
    \item We separate representation from supervision through controlled comparisons between standard and hypothesis-based evaluated baseline representations, both without and with matched ambiguity supervision, together with explicit evaluation on the very-high-ambiguity subset.
    \item We replace fold-level inference from one generated dataset with ten independent simulator datasets, group-disjoint splits, paired bootstrap intervals, two-sided tests, and Holm correction.
    \item We test the interpretation through shuffled-label, ambiguity-cutoff, margin, and leave-one-regime-out analyses; we additionally report frozen-model applicability to three-dimensionally validated real GUVs without recalibration.
\end{itemize}

\section{Related work}

\subsection{Selective prediction and uncertainty estimation}
The reject-option formulation dates to Chow's decision rule and the risk--coverage foundations of selective classification \citep{chow1970optimum,elyaniv2010foundations}. Later reject-option frameworks \citep{bartlett2008classification,cortes2016learning} and deep selective classifiers \citep{geifman2017selective,geifman2019selectivenet,liu2019deepgamblers,corbiere2019confidence} balance accuracy and coverage, with recent extensions for cost-sensitive learning, calibration, and post-hoc risk control \citep{charoenphakdee2021classification,narasimhan2024plugin,fisch2022calibrated,galil2023what}. Ensembles, dropout, evidential models, and distance-aware methods provide complementary epistemic-uncertainty scores \citep{gal2016dropout,lakshminarayanan2017deepensembles,sensoy2018evidential,charpentier2020posterior,liu2020simple,daxberger2021laplace}, including in sequential co-control \citep{huang2026probabilistic}. These approaches are strong baselines, but a scalar confidence score does not by itself identify whether uncertainty arises from ordinary prediction error, distributional mismatch, or structurally insufficient evidence \citep{hullermeier2021aleatoric}. CHASE instead represents the decision through competing explanations. Recent model-agnostic and post-hoc studies further show that strong selective performance can arise from classifier scores, cross-fitting, or training dynamics \citep{feng2022betterselective,pugnana2023modelagnostic,rabanser2022trainingdynamics}. CHASE is complementary to conformal risk control \citep{angelopoulos2021learn}, multi-threshold evaluation \citep{traub2024overcoming}, selective classification under shift \citep{liang2024selective}, and expert delegation \citep{narasimhan2022posthoc}.

\subsection{Multi-hypothesis prediction}
Multiple Choice Learning and related methods generate diverse plausible outputs \citep{guzman2012multiple,lee2016stochastic,lee2017confident}. Rupprecht-style multiple-hypothesis prediction \citep{rupprecht2017learning} and probabilistic U-Nets \citep{kohl2018probabilistic} explicitly represent ambiguity, but do not natively convert competition between hypotheses into a selective decision rule. CHASE uses the separation between class-conditional next-frame predictive losses as an abstention feature. To our knowledge, prior multi-hypothesis and selective-prediction methods have not explicitly evaluated collapse of this temporal likelihood margin as a structural-ambiguity signal under matched supervision.

\subsection{Temporal inference and early exiting}
Video models accumulate evidence over time \citep{furnari2021rolling,xu2019temporal,girdhar2021anticipative,wu2022memvit}, whereas early-exit systems primarily stop computation once a confidence criterion is met \citep{teerapittayanon2016branchynet,huang2018multi,kaya2019shallow,wu2019adaframe,ghodrati2021frameexit}. Some incorporate epistemic uncertainty \citep{bao2021evidential,guo2022uncertainty}, but their principal objective is computational efficiency rather than abstention caused by unresolved physical evidence. CHASE instead treats rejection as a safety-relevant decision about whether one temporal explanation is sufficiently better supported than its alternative.

\begin{figure}[pos=ht,width=\linewidth]
\centering
\includegraphics[width=\linewidth]{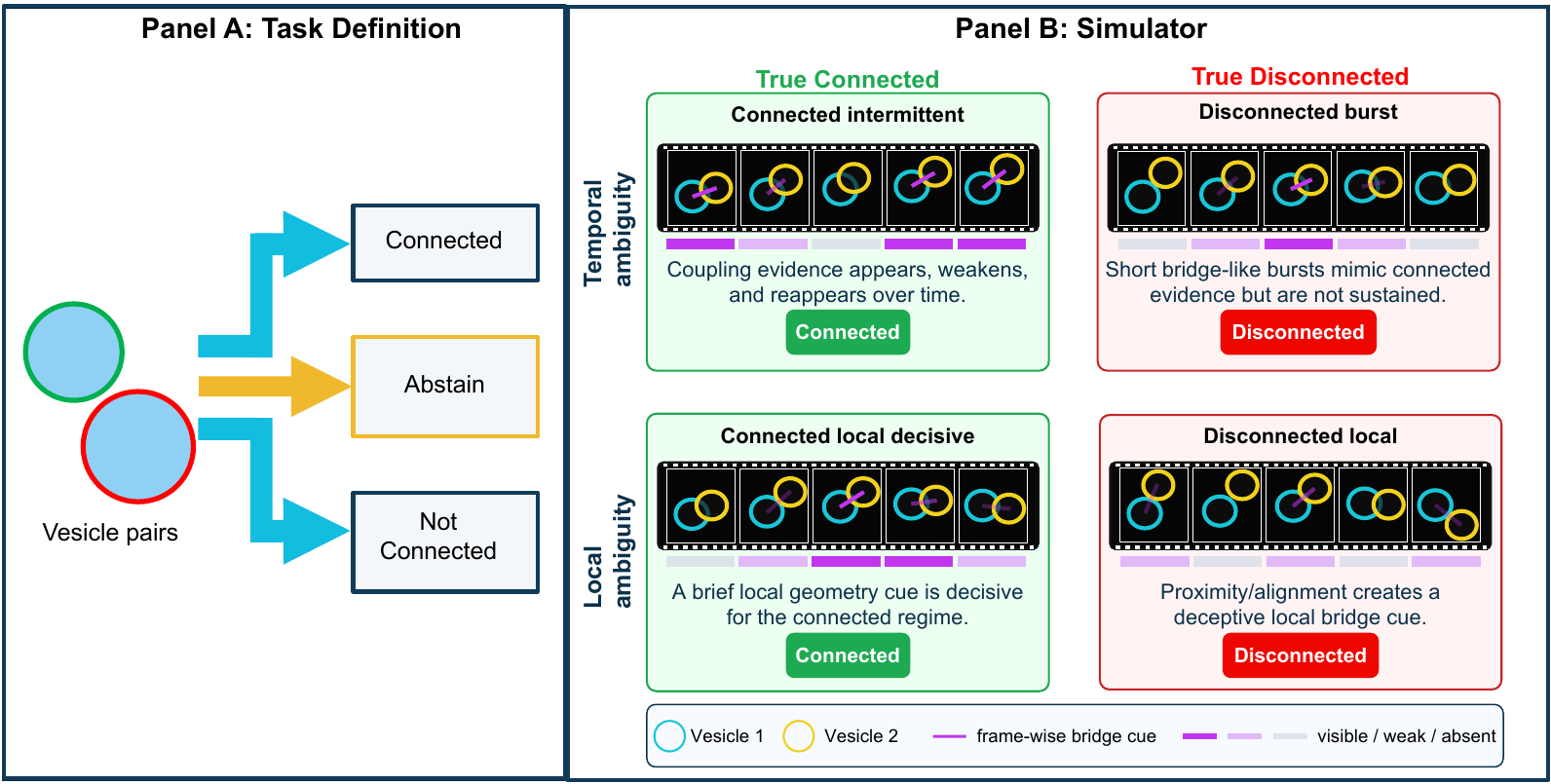}
\caption{\textbf{Task setup and simulator regimes.} The model classifies vesicle pairs as connected, disconnected, or abstained due to insufficient evidence. Using four simulated physical regimes of varying temporal and local ambiguity, we evaluate whether abstentions correlate with true ambiguity rather than transient noise.}
\label{fig:task}
\end{figure}
\FloatBarrier

\section{Problem Setup and Evaluation Metrics}
\label{sec:task}

\paragraph{Selective prediction under ambiguity}
Given a length-$T$ sequence $\mathbf{x}=\mathbf{x}_{1:T}=(\mathbf{x}_1,\ldots,\mathbf{x}_T)$, where each per-time-step descriptor $\mathbf{x}_t\in\mathbb{R}^{6}$, the baseline task is to predict the underlying connectivity state \(y\in\{\text{connected},\text{not\_connected}\}\). Because the connection evidence can be partially observable, some
sequences do not support a reliable binary decision. In the simulator, this is marked by an ambiguity label \(a\in\{0,1\}\) for each sequence, where \(a=1\) denotes structurally insufficient evidence. This binary ambiguity label \(a\) is not used to train the hypothesis backbone or auxiliary classifier and is never available at test time. It is used only to supervise the selector's accept/reject behavior in the simulated benchmark, so that abstention is trained against structural ambiguity rather than only binary prediction error. Thus, our simulated benchmark is an ambiguity-supervised selective prediction setting. The model outputs a committed
prediction $\hat{y}$, an accept score $s(\mathbf{x})$, and a final
three-way decision
$\hat{y}_{\mathrm{final}}\in\{\text{connected},\text{not\_connected},
\text{abstain}\}$. Abstention is triggered when
$s(\mathbf{x})<\tau$, where $\tau$ is chosen on validation data to meet
a target coverage $c\in\{0.80,0.90\}$ (see Figure~\ref{fig:task} panel A). Specifically, $\tau$ is defined as the empirical $(1-c)$-th quantile of validation acceptance scores $s(x)$, selecting the threshold that achieves the nearest attainable coverage to target $c$. Ties in score ranking are broken deterministically by dataset index, and in the edge case where an accepted set is empty, selective risk is conventionally defined as 0.

\paragraph{Evaluation metrics}
A useful selective predictor should keep errors low among accepted sequences while directing abstentions toward structurally ambiguous cases. For $n$ evaluation sequences, let $r_i = \mathbbm{1}[s(\mathbf{x}_i) < \tau]$ denote whether sequence $i$ is rejected (abstained). We report four primary metrics:
\begin{enumerate}[(1)]
    \item \textbf{No-abstain accuracy ($\NA$):} Standard binary classification accuracy when every sequence must be classified without rejection.
    \item \textbf{Risk at target coverage ($\Risk_c$):} The error rate computed exclusively among accepted sequences at a target coverage $c \in \{0.80, 0.90\}$: $\Risk_c = \frac{\sum_{i=1}^{n} (1 - r_i) \mathbbm{1}(\hat{y}_i \neq y_i)}{\sum_{i=1}^{n} (1 - r_i)}$.
    \item \textbf{Three-way accuracy ($\Three$):} Overall accuracy across all three possible decisions $\{\text{connected}, \text{not\_connected}, \text{abstain}\}$, where an ambiguous sequence ($a_i=1$) is correct only if rejected, and a non-ambiguous sequence ($a_i=0$) is correct only if committed to the true binary label: $\Three = \frac{1}{n} \sum_{i=1}^{n} \left[ a_i r_i + (1 - a_i)(1 - r_i) \mathbbm{1}(\hat{y}_i = y_i) \right]$.
    \item \textbf{Abstain alignment ($\Align$):} The precision of abstentions relative to true structural ambiguity: $\Align = \frac{\sum_{i=1}^{n} a_i r_i}{\sum_{i=1}^{n} r_i}$.
\end{enumerate}
On the very-high (VH) ambiguity subset, every sample is ambiguity-positive ($a_i=1$), making abstain alignment identically $100\%$ ($\Align = 100\%$); it is therefore omitted from VH comparative tables as non-discriminative.

\begin{figure}[pos=ht,width=\linewidth]
\centering
\includegraphics[width=\linewidth]{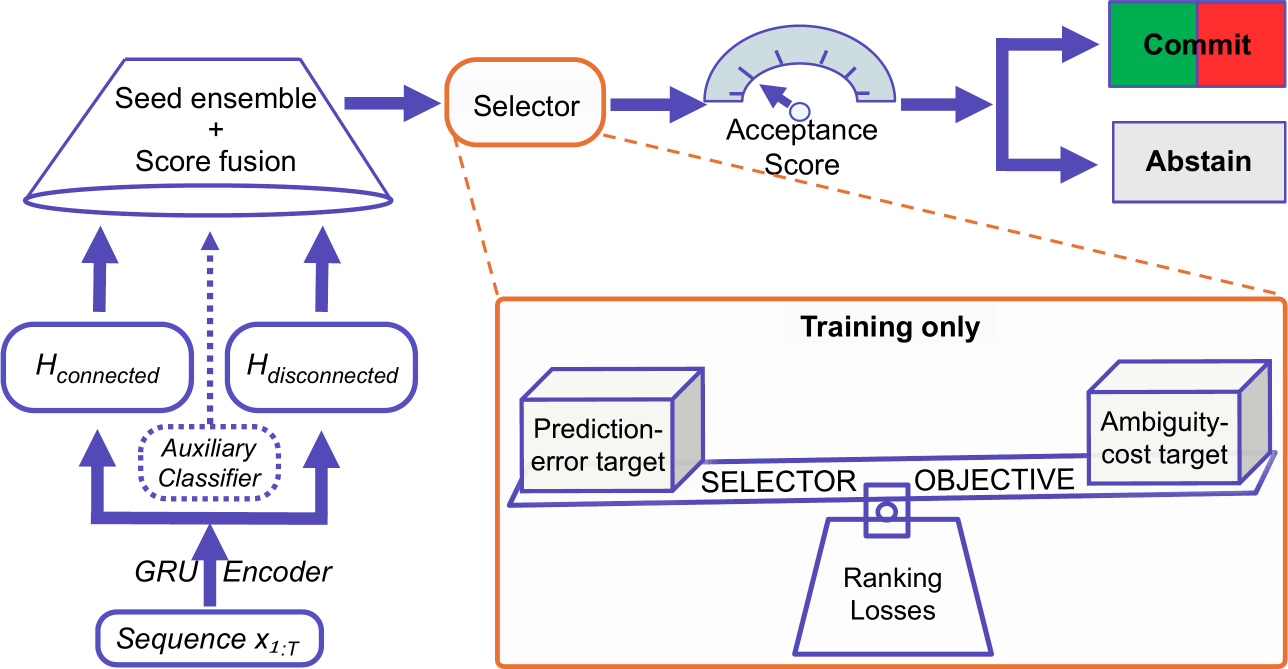}
\caption{\textbf{CHASE-Fusion pipeline and selector training.}
A shared GRU encoder feeds connected and disconnected predictive heads and
an auxiliary classifier. Seed-level outputs are ensembled and fused, and a
selector maps the resulting summaries to an acceptance score used to commit
the fused prediction or abstain. The inset summarizes training-only
supervision: binary cross-entropy on the ambiguity-aware rejection-cost target
together with pairwise ranking losses on the prediction-error and
ambiguity-cost targets. Ambiguity labels are not used at inference.}
\label{fig:method}
\end{figure}
\FloatBarrier

\section{Method}
\label{sec:method}

The pipeline (Figure~\ref{fig:method}) consists of three stages: (i) a \emph{hypothesis backbone} scoring candidate classes; (ii) a \emph{seed ensemble and decision fusion} capturing predictive disagreement; and (iii) a \emph{cost-aware pairwise selector} outputting the final accept score.

\subsection{Stage 1: Hypothesis backbone} The backbone follows a simple intuition: \emph{the right class should be the one whose learned dynamics predict the next frame more accurately.} We instantiate two predictive heads---one for each class---and let them compete. A shared Gated Recurrent Unit (GRU) \citep{cho2014learning} encodes context $x_{1:t}$ into a hidden state $h_t$. Two Gaussian heads (for $c=\text{connected}$ and $n=\text{not\_connected}$) emit a mean and log-variance from $h_t$ to predict the next frame $x_{t+1}$. The frame-level negative log-likelihoods are time-averaged to yield per-class scores $\ell^{c}$ and $\ell^{n}$. Simultaneously, an auxiliary head on pooled $h_{1:T-1}$ is trained on the standard binary classification task to output a baseline probability $\pi_{\mathrm{aux}}$. The auxiliary head provides an additional discriminative probability for the fused CHASE-Fusion configuration. With a fixed class-separation margin $m\geq 0$, margin-loss weight $\lambda_m$, and auxiliary-classification weight $\lambda_c$, the backbone optimizes:
\begin{equation}
\label{eq:backbone-loss}
\mathcal{L}_{\text{backbone}} \;=\; \ell^{y} \;+\;\lambda_{m}\big[m-(\ell^{\bar y}-\ell^{y})\big]_{+} \;+\;\lambda_{c}\mathrm{CE}(\pi^{\text{aux}},y),
\end{equation}
where $\bar{y}$ is the incorrect class and
$[z]_+=\max(0,z)$. This forces the correct hypothesis to explain the data while maintaining a margin between hypotheses---a gap we exploit to detect ambiguity. We extract the hypothesis-margin softmax $\pi^{\text{hyp}} = \mathrm{softmax}(-\ell^{c},\,-\ell^{n})$ and $\pi^{\text{aux}}$ for the selector.

\subsection{Stage 2: Seed ensemble and decision fusion} To reduce sensitivity to random initialization, minibatch ordering, and
stochastic optimization, we train \(K=3\) backbones with different random
seeds. Per-sequence outputs are averaged to obtain ensemble versions of $\pi^{\text{hyp}}$, $\pi^{\text{aux}}$, and the score gap. The cross-seed dispersion---specifically, the standard deviations $\sigma^{\text{hyp}}$ and $\sigma^{\text{aux}}$, and the fraction of seeds disagreeing with the majority ($\delta$)---feeds directly into the selector's feature space. We fuse the probability sources via a scalar $\beta \in [0,1]$, selected from
\(\{0,0.05,\ldots,1\}\) using the corresponding validation partition: $\pi^{\text{fused}} = \beta\pi^{\text{hyp}} + (1-\beta)\pi^{\text{aux}}$. The committed prediction is $\hat{y} = \arg\max \pi^{\text{fused}}$.

\label{sec:selector}
\subsection{Stage 3: Cost-aware pairwise selector} We train a lightweight MLP $f_{\theta}$ to map ensemble-fused summaries and dispersion signals $\phi(\mathbf{x})$ to a scalar rejection-cost logit. An error-only target $E(\mathbf{x})=\mathbbm{1}[\hat{y}\neq y]$ does not distinguish an unambiguous mistake from a correct but structurally unresolved case. We therefore define an \emph{ambiguity-aware rejection-cost target}
$y^{\text{cost}}(\mathbf{x})=\max(E(\mathbf{x}),\gamma a)\in[0,1]$.
Errors carry full cost, ambiguous-but-correct sequences carry cost $\gamma$, and ambiguous-and-wrong sequences carry full cost. The selector optimizes binary cross-entropy together with two pairwise ranking losses:
\begin{equation}
\label{eq:selector-loss}
\mathcal{L}_{\text{sel}}=\mathrm{BCE}\!\big(f_{\theta}(\phi),y^{\text{cost}}\big)
+\lambda_E\mathcal{L}_{\text{rank}}(E)
+\lambda_A\mathcal{L}_{\text{rank}}(y^{\text{cost}}).
\end{equation}
Here, $\mathcal{L}_{\text{rank}}(z)$ is a softplus margin loss applied to pairs $(i,j)$ with $z_i>z_j$, weighted by $z_i-z_j$. The error-ranking term prioritizes low risk, whereas the ambiguity-cost term ranks structurally unresolved sequences toward rejection when the coverage budget permits. The final accept score is $s(\mathbf{x})=1-\sigma(f_{\theta}(\phi(\mathbf{x})))$, and the model commits when $s(\mathbf{x})\geq\tau$. The validation-selected weights are $\gamma=0.62$, $\lambda_E=0.66$, and $\lambda_A=0.10$; the hyperparameter sweep is retained in Supplementary Table~S4.

\subsection{Controlled comparison variants}
\label{sec:controlled_variants}
We distinguish the hypothesis representation from the supervision used by the selector. \textbf{S-E} uses a standard three-member single-GRU classifier ensemble with an error-only selector, whereas \textbf{H-E} uses the three-member competing-hypothesis ensemble with the same error-only target. \textbf{S-A} and \textbf{H-A} retain these respective representations but train the common selector with the identical simulator ambiguity target. The auxiliary head provides a multitask classification loss during backbone training. Its outputs are integrated into the committed prediction and selector features only in CHASE-Fusion; the H-E and H-A
controls retain the auxiliary training loss, but their committed predictions and selector inputs use only hypothesis-derived outputs.  The \textbf{raw hypothesis margin} predicts from the sign of $\ell^{n}-\ell^{c}$ and abstains when the absolute margin falls below a validation-selected threshold. \textbf{CHASE-Fusion} combines the hypothesis-derived and auxiliary-classifier summaries before selection. These variants provide controlled comparisons of hypothesis representation and selector supervision, while the H-A-to-CHASE-Fusion comparison evaluates the incremental value of auxiliary-output integration. Full-pipeline performance claims refer to CHASE-Fusion; H-A-versus-S-A claims refer specifically to the controlled representation comparison.

\section{GUV-inspired simulator and real-data scope}
\label{sec:data}

\paragraph{\textbf{GUV-Inspired Simulator Design.}}
We developed a targeted 2D kinematic simulator to systematically stress-test selective prediction under controlled ambiguity. The vesicles move in a unit box under Brownian motion, optional shared drift, and---for connected pairs---spring coupling that keeps them physically linked. As illustrated in Figure~\ref{fig:task} panel B, the simulator generates 64-frame sequences (approximately 6~s) of vesicle pairs under two latent states: connected and disconnected. 

\paragraph{\textbf{Latent activity profiles and matched pairs.}} To ensure classifiers rely on temporal reasoning rather than simple global summaries, the simulator uses \emph{latent activity profiles} $u(t) \in [0,1]$ that control when evidence becomes visible. For connected sequences, $u(t)$ modulates the strength of the spring coupling or the visibility of the visual ``bridge'' (representing the physical membrane neck). For matched negative sequences, the identical $u(t)$ timing is reused, but instead of physical coupling, it triggers shared fluid drift and temporary proximity. This creates \emph{matched pairs} with similar global summary statistics but different causal mechanisms. Matched pairs generated from the same latent activity profile were assigned to the same train, validation, or test partition to prevent profile-level leakage. We focus on two challenging regimes (Figure~\ref{fig:task} panel B): 
(1) \textit{Intermittent coupling:} Dynamic spring linkage is matched with a negative sequence exhibiting coordinated shared drift. 
(2) \textit{Short local evidence:} A brief window of strong coupling is matched with a brief motion distractor. 
These pairings force the model to distinguish true physical linkage from coincidental proximity.

\paragraph{\textbf{Features, ambiguity, and dataset splits.}} For each frame, we extract six continuous features that form the input vector $\mathbf{x}$ (as defined in Section~\ref{sec:task}): Euclidean distance (between vesicle centroids), distance change (frame-to-frame), relative motion (alignment of velocity vectors), spatial support count (bridge region pixel area), median bridge score (visual intensity/confidence), and median bridge width. Sequences are assigned an ambiguity level $\alpha \in [0,1]$, which scales observation noise and degrades visual bridge evidence. Sequences with $\alpha \geq 0.75$ are explicitly flagged as \texttt{truly\_ambiguous} ($a=1$) and constitute the very-high (VH) ambiguity subset, which we contrast against the overall dataset (full test pool).

\paragraph{\textbf{Frozen-model real-GUV applicability protocol.}} Real phase-separated GUVs were subjected to hypertonic stress and imaged via confocal microscopy, with full protocols in Supplementary Section~S5. For the pilot cohort summarized in Section~\ref{sec:real_guv_results} and Supplementary Table~S3, physical connectivity was assessed using three-dimensional reconstruction. The available three-dimensional reconstructions establish physical attachment, but the fine neck can remain unresolved in individual projections. Thus, these real videos provide qualitative context, but not an extensive quantitative real-data benchmark. Real 2D projections of 3D physical processes can produce misleading visual evidence: connected pairs may have weak or out-of-plane necks, while visually disconnected pairs may show correlated motion due to drift. These data encapsulate the partial observability problem. 

We evaluated the frozen simulator-trained pipeline on 11 pair-focused sparse-frame real-GUV cases with three-dimensionally established physical connectivity labels: ten connected cases and one disconnected case. The experimental constraints underlying this cohort composition
are discussed in Section~\ref{sec:discussion}. A fixed set of frames per video case was selected during preprocessing to preserve vesicle-pair identity and obtain usable geometric segmentations. Frame selection preceded CHASE scoring and was not subsequently altered on the basis of acceptance scores or final predictions; all selected frames were retained in the reported analysis. Each selected frame was converted, without access to its physical label, into a simulator-style geometric representation from which the same six temporal descriptors used by CHASE were computed. The resulting descriptor sequence was interpolated to the 64-step input length expected by the frozen model. The model, fusion rule, selector, and simulator-validation thresholds remained unchanged. The primary analysis used the strict canonical representation and the simulator-derived nominal 80\% operating point, with the final decision obtained by consensus across the five frozen simulator folds. These five frozen fold models are used only for the real-GUV applicability pilot; the principal inferential benchmark instead uses ten independently generated simulator datasets, with simulator seed as the independent unit.

\section{Experimental design and statistical analysis}
\label{sec:experimental_design}

\subsection{Independent simulator replication}
We generate ten independent simulator realizations using simulator seeds 202601--202610. Using independently generated simulator datasets rather than repeated
folds on a single realization evaluates the reproducibility of
benchmark performance across stochastic simulator instantiations. Each dataset contains 3,360 sequences grouped into 1,680 matched positive--negative pairs. For each simulator seed, a single group-disjoint split assigns 1,075 groups (2,150 sequences) to training, 269 groups (538 sequences) to validation, and 336 groups (672 sequences) to testing. Stratification uses source partition, ambiguity regime, and ambiguity bin; paired sequences derived from the same latent activity profile never cross partitions. Simulator seed, rather than fold, is the independent experimental unit. The simulator configuration, architecture, training schedule, and selector hyperparameters were fixed before evaluating these ten seeds and remain unchanged across seeds.

\subsection{Baselines and supervision parity}
We compare CHASE against a standard single-branch Maximum Softmax Probability (MSP) baseline, alongside canonical methods from two families: epistemic uncertainty estimation---MC Dropout \citep{gal2016dropout} and Deep Ensembles \citep{lakshminarayanan2017deepensembles}---and learned selective classification---SelectiveNet \citep{geifman2019selectivenet}, Deep Gamblers \citep{liu2019deepgamblers}, ConfidNet \citep{corbiere2019confidence}, and a five-head Rupprecht-style multiple-hypothesis prediction (MHP) baseline \citep{rupprecht2017learning} trained with the relaxed winner-takes-all objective and evaluated with the variance-based acceptance score; we denote this fixed implementation MHP-var. Beyond the included Rupprecht-style MHP baseline, we do not add further multi-output models lacking a native selective decision rule; temporal early-exit networks are also outside the comparison because they are optimized primarily for computational efficiency rather than ambiguity-aware abstention. For comparison, all external baselines use the same precomputed six-feature input and are adapted to a GRU backbone, reducing input-representation confounding while comparing their uncertainty and abstention mechanisms. No ambiguity label is available to the predictive backbone or at test time.

\subsection{Implementation details}

The CHASE and matched-GRU implementations used Python~3.10.19 and
64-step sequences with six features per step. Input features were
standardized using statistics estimated only from the corresponding
fitting partition. Models were optimized with Adam using a batch size of
64 and a learning rate of \(10^{-3}\). The standard single-branch GRU and
the hypothesis backbone both used hidden dimension 64. The standard GRU
was trained for at most 40 epochs, whereas the hypothesis backbone was
trained for at most 30 epochs; both used early-stopping patience 6. The hypothesis-backbone objective used a class-separation margin \(m=1\),
fixed before evaluation on the ten independent simulator realizations,
together with \(\lambda_m=1\) and \(\lambda_c=1.5\).

Each CHASE ensemble contained three backbone members initialized with
seeds 42, 143, and 244. For CHASE-Fusion, the fusion coefficient was
selected separately for each validation split from
\(\{0,0.05,\ldots,1\}\) by maximizing forced binary validation accuracy,
with lower validation negative log-likelihood used to break accuracy
ties. The selected coefficient was fixed during test evaluation. H-E and
H-A used \(\beta=1\).

The outer validation scores were divided group-disjointly into 65\%
selector-fitting and 35\% selector-tuning subsets. The selector was an MLP
with two hidden layers of width 24 and dropout 0.10. It was optimized with
Adam at learning rate \(10^{-3}\) for at most 80 epochs with patience 8.
The ambiguity cost and ranking-loss weights were
\(\gamma=0.62\), \(\lambda_E=0.66\), and \(\lambda_A=0.10\).
At most 4096 ordered pairs were sampled per ranking-loss evaluation, with
ranking margin zero. The development sweep for the three selector weights
is reported in Supplementary Table~S4.

% \subsection{Implementation details}

% The CHASE and matched-GRU implementations used Python~3.10.19. Inputs comprised 64 time steps with
% six features. Models were trained with Adam using a batch size of 64 and
% a learning rate of $10^{-3}$. The standard GRU and hypothesis backbone
% used hidden dimension 64, maximum training lengths of 40, and early-stopping patience 6. The CHASE ensemble used
% three backbone seeds (42, 143, and 244). The selector contained two
% hidden layers of width 24 with dropout 0.10, was trained for at most
% 80 epochs with patience 8, and sampled at most 4096 ranking pairs.

\subsection{Metrics and inference}
\label{sec:metrics_and_inference}
Evaluations report the metrics defined in Section~\ref{sec:task} ($\NA$, $\Risk_c$, $\Three$, and $\Align$), alongside achieved coverage, and ambiguity AUROC. Summary tables present the mean $\pm$ sample standard deviation across the ten independent simulator seeds. 

Primary paired comparisons treat the ten simulator seeds as the paired resampling and testing units. Statistical significance is assessed using 50,000-resample paired bootstrap confidence intervals, exact two-sided sign-flip tests, and two-sided Wilcoxon signed-rank tests. To control family-wise error rates, Holm--Bonferroni adjustment is applied within each metric--coverage--subset family, with significance claimed at adjusted $p < 0.05$.

\subsection{Unseen-regime evaluation}
We additionally train on one ambiguity mechanism and test on the other: intermittent coupling $\rightarrow$ local evidence, and local evidence $\rightarrow$ intermittent coupling. Unlike the primary independent-seed protocol, this experiment first partitions the simulator by ambiguity mechanism and therefore uses a separate train/validation/test split within each regime. Each direction uses 1,200 training, 240 validation, and 240 test sequences per simulator seed. Because validation thresholds need not preserve nominal coverage under
shift, we report achieved coverage and accepted-set risk at the
transferred threshold, together with standardized risk at 80\%
coverage. Complete risk--coverage curves for both leave-one-regime-out
directions are provided in Supplementary Figure~S2.

\begin{table*}[pos=t,width=\textwidth]
\centering
\caption{\textbf{Independent-seed benchmark against established selective-prediction baselines.} Performance across forced prediction (Panel a), 90\% target coverage (Panel b), and 80\% target coverage (Panel c) on independent simulator seeds. MHP-var denotes the single five-head, variance-score Rupprecht-style MHP implementation.}
\label{tab:literature_benchmark}
\small
\renewcommand{\arraystretch}{1.04}
\setlength{\tabcolsep}{2pt}

\textit{Panel (a): Forced prediction.}
\begin{tabular*}{\textwidth}{@{\extracolsep{\fill}}lcc@{}}
\toprule
Method & $\NA$(O) & $\NA$(V) \\
\midrule
MSP & $90.06\pm0.61$ & $86.51\pm2.49$ \\
MC Dropout & $89.43\pm0.75$ & $85.06\pm3.48$ \\
Deep Ensemble & $90.52\pm0.70$ & \best{88.14\pm2.95} \\
SelectiveNet & $88.30\pm1.67$ & $83.26\pm2.83$ \\
Deep Gamblers & $87.35\pm0.86$ & $82.24\pm2.79$ \\
ConfidNet & $90.06\pm0.61$ & $86.51\pm2.49$ \\
MHP-var & $85.15\pm1.16$ & $77.71\pm3.79$ \\
CHASE-Fusion & \best{91.01\pm0.51} & $87.02\pm3.05$ \\
\bottomrule
\end{tabular*}

\vspace{0.55em}
\textit{Panel (b): Target coverage 90\%.}
\begin{tabular*}{\textwidth}{@{\extracolsep{\fill}}lccccc@{}}
\toprule
Method & $\Risk_{90}$(O) & $\Risk_{90}$(V) & $\Three_{90}$(O) & $\Three_{90}$(V) & $\Align_{90}$(O) \\
\midrule
MSP & $6.45\pm0.54$ & $9.06\pm2.45$ & $67.43\pm1.28$ & $13.24\pm2.61$ & $33.05\pm3.58$ \\
MC Dropout & $6.71\pm1.00$ & $9.86\pm3.72$ & $68.05\pm1.40$ & $14.57\pm2.28$ & $36.61\pm5.28$ \\
Deep Ensemble & \best{5.96\pm0.98} & $8.03\pm3.01$ & $67.72\pm1.66$ & $13.55\pm2.06$ & $33.91\pm5.28$ \\
SelectiveNet & $8.74\pm1.64$ & $12.38\pm4.72$ & $68.35\pm2.35$ & \best{18.05\pm6.31} & $43.97\pm10.89$ \\
Deep Gamblers & $9.63\pm1.08$ & $12.46\pm4.37$ & $66.88\pm1.95$ & $17.32\pm4.37$ & $41.52\pm6.07$ \\
ConfidNet & $6.63\pm1.08$ & $9.37\pm2.26$ & $67.34\pm1.51$ & $13.58\pm1.98$ & $33.82\pm6.58$ \\
MHP-var & $13.92\pm2.28$ & $21.81\pm5.31$ & $63.56\pm3.72$ & $14.02\pm5.34$ & $33.61\pm11.82$ \\
CHASE-Fusion & $6.04\pm0.70$ & \best{7.87\pm2.62} & \best{70.25\pm2.04} & $16.15\pm1.94$ & \best{47.61\pm7.27} \\
\bottomrule
\end{tabular*}

\vspace{0.55em}
\textit{Panel (c): Target coverage 80\%.}
\begin{tabular*}{\textwidth}{@{\extracolsep{\fill}}lccccc@{}}
\toprule
Method & $\Risk_{80}$(O) & $\Risk_{80}$(V) & $\Three_{80}$(O) & $\Three_{80}$(V) & $\Align_{80}$(O) \\
\midrule
MSP & $4.40\pm0.74$ & $5.91\pm2.32$ & $65.97\pm1.31$ & $26.04\pm3.72$ & $33.90\pm5.02$ \\
MC Dropout & $4.50\pm1.17$ & $6.65\pm4.31$ & $66.21\pm1.16$ & $26.82\pm3.17$ & $34.49\pm3.54$ \\
Deep Ensemble & \best{3.82\pm0.77} & $5.08\pm2.75$ & $66.58\pm1.49$ & $27.52\pm3.04$ & $35.06\pm3.97$ \\
SelectiveNet & $6.62\pm1.83$ & $9.17\pm4.54$ & $67.14\pm2.88$ & $32.08\pm6.37$ & $40.17\pm4.70$ \\
Deep Gamblers & $6.86\pm1.17$ & $9.81\pm3.48$ & $65.36\pm1.60$ & $29.16\pm4.26$ & $36.16\pm3.02$ \\
ConfidNet & $4.51\pm1.13$ & $6.44\pm2.74$ & $66.00\pm1.69$ & $26.13\pm2.09$ & $34.15\pm4.17$ \\
MHP-var & $12.81\pm3.41$ & $21.61\pm6.19$ & $61.74\pm5.28$ & $28.56\pm8.23$ & $33.95\pm8.22$ \\
CHASE-Fusion & $4.25\pm0.73$ & \best{4.81\pm2.33} & \best{71.18\pm2.01} & \best{34.97\pm4.42} & \best{48.12\pm5.91} \\
\bottomrule
\end{tabular*}

\vspace{0.4em}
\parbox{\textwidth}{\footnotesize \textit{Note:} Values are percentages, reported as mean $\pm$ sample standard deviation over ten independent simulator datasets. Boldface identifies the best mean within each column and is descriptive; it does not by itself indicate a statistically significant difference. O is the complete test set and V is the very-high-ambiguity subset. $\Align$(V) is $100.00\pm0.00$ for every method because every V sample is ambiguity-positive; it is omitted as non-discriminative. This note is also applicable for Table \ref{tab:no_ambiguity_complete} and \ref{tab:matched_ambiguity_complete}.}
\end{table*}

\section{Results}
\label{sec:results}

\subsection{Independent comparison with established baselines}
\label{sec:baseline_comparison}
Table~\ref{tab:literature_benchmark} shows comparison with the baselines while exposing the central tradeoff. At 80\% coverage, CHASE-Fusion improves overall three-way accuracy by 4.04 percentage points (pp) and abstain alignment by 7.95 pp over the best non-CHASE means, while its risk is 0.43 pp above Deep Ensemble. At 90\%, the corresponding gains are 1.90 and 3.64 pp, and the risk gap narrows to 0.08 pp. The independent replication therefore supports stronger ambiguity-aware decisions at a competitive, rather than uniformly superior, risk boundary.

The VH subset makes this distinction sharper. At 80\%, CHASE-Fusion reduces mean VH risk by 0.27 pp relative to Deep Ensemble and improves VH three-way accuracy by 2.89 pp over SelectiveNet. At 90\%, it retains the lowest VH risk by 0.16 pp relative to Deep Ensemble, whereas SelectiveNet has 1.90 pp higher VH three-way accuracy. Thus, the advantage is strongest when sufficient rejection budget remains to target structurally ambiguous cases. The benchmark table here provides an overall comparison with established methods, whereas formal paired inference is reserved for the prespecified controlled representation and supervision comparisons that address the paper's primary mechanistic hypotheses. Complete in-distribution risk--coverage curves for the overall test set and the very-high-ambiguity subset are provided in Supplementary Figure~~S1.

\begin{table*}[pos=t,width=\textwidth]
\centering
\caption{\textbf{Complete comparison without ambiguity supervision.} Performance across forced prediction (Panel a), 90\% target coverage (Panel b), and 80\% target coverage (Panel c). S-E and H-E use the identical error-only selector on standard-ensemble and hypothesis representations. The raw margin uses neither a learned selector nor the simulator ambiguity target.}
\label{tab:no_ambiguity_complete}
\small
\renewcommand{\arraystretch}{1.04}
\setlength{\tabcolsep}{2pt}

\textit{Panel (a): Forced prediction.}
\begin{tabular*}{\textwidth}{@{\extracolsep{\fill}}lcc@{}}
\toprule
Method & $\NA$(O) & $\NA$(V) \\
\midrule
MSP & $90.06\pm0.61$ & $86.51\pm2.49$ \\
MC Dropout & $89.43\pm0.75$ & $85.06\pm3.48$ \\
Deep Ensemble & $90.52\pm0.70$ & \best{88.14\pm2.95} \\
SelectiveNet & $88.30\pm1.67$ & $83.26\pm2.83$ \\
Deep Gamblers & $87.35\pm0.86$ & $82.24\pm2.79$ \\
ConfidNet & $90.06\pm0.61$ & $86.51\pm2.49$ \\
MHP-var & $85.15\pm1.16$ & $77.71\pm3.79$ \\
S-E & $90.13\pm0.73$ & $86.41\pm2.41$ \\
Raw margin & \best{91.15\pm0.69} & $87.16\pm3.01$ \\
H-E & \best{91.15\pm0.69} & $87.16\pm3.01$ \\
\bottomrule
\end{tabular*}

\vspace{0.55em}
\textit{Panel (b): Target coverage 90\%.}
\begin{tabular*}{\textwidth}{@{\extracolsep{\fill}}lccccc@{}}
\toprule
Method & $\Risk_{90}$(O) & $\Risk_{90}$(V) & $\Three_{90}$(O) & $\Three_{90}$(V) & $\Align_{90}$(O) \\
\midrule
MSP & $6.45\pm0.54$ & $9.06\pm2.45$ & $67.43\pm1.28$ & $13.24\pm2.61$ & $33.05\pm3.58$ \\
MC Dropout & $6.71\pm1.00$ & $9.86\pm3.72$ & $68.05\pm1.40$ & $14.57\pm2.28$ & $36.61\pm5.28$ \\
Deep Ensemble & $5.96\pm0.98$ & $8.03\pm3.01$ & $67.72\pm1.66$ & $13.55\pm2.06$ & $33.91\pm5.28$ \\
SelectiveNet & $8.74\pm1.64$ & $12.38\pm4.72$ & $68.35\pm2.35$ & \best{18.05\pm6.31} & $43.97\pm10.89$ \\
Deep Gamblers & $9.63\pm1.08$ & $12.46\pm4.37$ & $66.88\pm1.95$ & $17.32\pm4.37$ & $41.52\pm6.07$ \\
ConfidNet & $6.63\pm1.08$ & $9.37\pm2.26$ & $67.34\pm1.51$ & $13.58\pm1.98$ & $33.82\pm6.58$ \\
MHP-var & $13.92\pm2.28$ & $21.81\pm5.31$ & $63.56\pm3.72$ & $14.02\pm5.34$ & $33.61\pm11.82$ \\
S-E & $6.89\pm0.78$ & $9.83\pm3.01$ & $67.69\pm1.22$ & $13.50\pm1.95$ & $35.37\pm4.40$ \\
Raw margin & \best{5.74\pm0.87} & $8.38\pm2.96$ & $69.75\pm1.79$ & $16.55\pm4.47$ & $42.32\pm5.31$ \\
H-E & $5.85\pm0.68$ & \best{8.00\pm3.22} & \best{69.87\pm2.55} & $16.13\pm3.08$ & \best{44.36\pm8.38} \\
\bottomrule
\end{tabular*}

\vspace{0.55em}
\textit{Panel (c): Target coverage 80\%.}
\begin{tabular*}{\textwidth}{@{\extracolsep{\fill}}lccccc@{}}
\toprule
Method & $\Risk_{80}$(O) & $\Risk_{80}$(V) & $\Three_{80}$(O) & $\Three_{80}$(V) & $\Align_{80}$(O) \\
\midrule
MSP & $4.40\pm0.74$ & $5.91\pm2.32$ & $65.97\pm1.31$ & $26.04\pm3.72$ & $33.90\pm5.02$ \\
MC Dropout & $4.50\pm1.17$ & $6.65\pm4.31$ & $66.21\pm1.16$ & $26.82\pm3.17$ & $34.49\pm3.54$ \\
Deep Ensemble & $3.82\pm0.77$ & $5.08\pm2.75$ & $66.58\pm1.49$ & $27.52\pm3.04$ & $35.06\pm3.97$ \\
SelectiveNet & $6.62\pm1.83$ & $9.17\pm4.54$ & $67.14\pm2.88$ & \best{32.08\pm6.37} & $40.17\pm4.70$ \\
Deep Gamblers & $6.86\pm1.17$ & $9.81\pm3.48$ & $65.36\pm1.60$ & $29.16\pm4.26$ & $36.16\pm3.02$ \\
ConfidNet & $4.51\pm1.13$ & $6.44\pm2.74$ & $66.00\pm1.69$ & $26.13\pm2.09$ & $34.15\pm4.17$ \\
MHP-var & $12.81\pm3.41$ & $21.61\pm6.19$ & $61.74\pm5.28$ & $28.56\pm8.23$ & $33.95\pm8.22$ \\
S-E & $4.39\pm0.78$ & $6.50\pm3.12$ & $66.38\pm1.32$ & $27.10\pm3.43$ & $35.15\pm5.31$ \\
Raw margin & \best{3.80\pm0.82} & \best{4.91\pm1.86} & $68.48\pm2.01$ & $28.25\pm5.04$ & $39.09\pm4.98$ \\
H-E & $4.03\pm0.74$ & $4.96\pm2.60$ & \best{68.91\pm3.77} & $29.98\pm5.75$ & \best{41.56\pm9.92} \\
\bottomrule
\end{tabular*}
\end{table*}

\subsection{Hypothesis comparison without ambiguity supervision}
Table~\ref{tab:no_ambiguity_complete} isolates the representation effect without the simulator ambiguity target. The raw margin and Deep Ensemble have essentially identical overall risk at 80\% coverage. Nevertheless, the raw margin improves overall three-way accuracy by 1.90 pp (95\% paired-bootstrap CI 1.04--2.81) and alignment by 4.03 pp (CI 1.53--6.64); the Holm-adjusted tests remain significant. At 90\%, the gains are approximately 2.0 and 8.4 pp, again without a significant risk difference.

The VH results are directionally consistent but more variable. Relative to Deep Ensemble, the raw margin changes VH risk by $-0.17$ pp and VH three-way accuracy by $+0.7$ pp at 80\%, and by $+0.35$ and $+3.00$ pp at 90\%; these VH differences are not significant after correction. The learned H-E selector is more consistently favorable than S-E on VH cases, lowering mean risk by 1.54 pp and improving three-way accuracy by 2.88 pp at 80\%, with corresponding changes of $-1.83$ and $+2.63$ pp at 90\%. The conclusion without ambiguity supervision is therefore specific: hypothesis comparison preserves the low-risk boundary while improving how the limited abstention budget is allocated.

\subsection{Hypothesis representations benefit under matched ambiguity supervision}
Table~\ref{tab:matched_ambiguity_complete} gives each representation the same ambiguity-supervised selector. On the overall test set, H-A improves three-way accuracy over S-A by 9.75 pp (CI 8.59--11.00) and alignment by 28.3 pp (CI 24.63--32.10) at 80\%, with no significant risk difference. At 90\%, the gains remain 3.74 and 20.25 pp. H-A also exceeds the strongest external ambiguity-supervised baseline by 8.44 pp in three-way accuracy and 18.56 pp in alignment at 80\%; its overall risk is 0.23 pp above the matched S-A comparator (4.58\% versus 4.35\%).

The VH comparison provides the clearest evidence that the representation ranks structural ambiguity rather than merely ordinary error. At 80\%, H-A lowers VH risk by 3.46 pp and improves VH three-way accuracy by 19.1 pp relative to S-A; both effects have the same direction on all ten seeds and remain significant after Holm correction. At 90\%, the mean changes are $-1.31$ and $+5.8$ pp, but the three-way difference narrowly misses the corrected threshold. H-A also improves over CHASE-Fusion by 5.00 pp in overall three-way accuracy and 15.3 pp in alignment at 80\%, with no significant risk difference. Accordingly, H-A is the cleanest test of the hypothesis representation, whereas CHASE-Fusion is retained as the original fusion-based operating point. Complete in-distribution risk--coverage curves for the overall and
very-high-ambiguity subsets are provided in Supplementary
Figure~S1; they complement the fixed-coverage results in
Tables~\ref{tab:no_ambiguity_complete} and
\ref{tab:matched_ambiguity_complete}.

\begin{table*}[pos=t,width=\textwidth]
\centering
\caption{\textbf{Complete matched ambiguity-supervision comparison.} Performance across forced prediction (Panel a), 90\% target coverage (Panel b), and 80\% target coverage (Panel c). All methods marked +a, as well as S-A and H-A, use the identical simulator ambiguity target in the common selector, controlling access to ambiguity supervision while comparing the evaluated predictive representations.  MHP+$a$ uses the same five-head MHP predictive representation as MHP-var, but the common ambiguity-supervised selector replaces the native variance-based acceptance score.}
\label{tab:matched_ambiguity_complete}
\small
\renewcommand{\arraystretch}{1.04}
\setlength{\tabcolsep}{2pt}

\textit{Panel (a): Forced prediction.}
\begin{tabular*}{\textwidth}{@{\extracolsep{\fill}}lcc@{}}
\toprule
Method & $\NA$(O) & $\NA$(V) \\
\midrule
MSP+$a$ & $90.06\pm0.61$ & $86.51\pm2.49$ \\
MC Dropout+$a$ & $89.43\pm0.75$ & $85.06\pm3.48$ \\
S-A & $90.13\pm0.73$ & $86.41\pm2.41$ \\
SelectiveNet+$a$ & $88.30\pm1.67$ & $83.26\pm2.83$ \\
Deep Gamblers+$a$ & $87.35\pm0.86$ & $82.24\pm2.79$ \\
ConfidNet+$a$ & $90.06\pm0.61$ & $86.51\pm2.49$ \\
MHP+$a$ & $85.15\pm1.16$ & $77.71\pm3.79$ \\
CHASE-Fusion & $91.01\pm0.51$ & $87.02\pm3.05$ \\
H-A & \best{91.15\pm0.69} & \best{87.16\pm3.01} \\
\bottomrule
\end{tabular*}

\vspace{0.55em}
\textit{Panel (b): Target coverage 90\%.}
\begin{tabular*}{\textwidth}{@{\extracolsep{\fill}}lccccc@{}}
\toprule
Method & $\Risk_{90}$(O) & $\Risk_{90}$(V) & $\Three_{90}$(O) & $\Three_{90}$(V) & $\Align_{90}$(O) \\
\midrule
MSP+$a$ & $6.55\pm0.59$ & $9.32\pm2.72$ & $67.46\pm1.35$ & $12.74\pm2.68$ & $32.80\pm3.64$ \\
MC Dropout+$a$ & $7.25\pm0.85$ & $10.54\pm4.43$ & $67.65\pm1.36$ & $14.50\pm3.24$ & $35.91\pm5.74$ \\
S-A & $6.53\pm0.77$ & $9.28\pm3.22$ & $67.87\pm1.24$ & $13.65\pm2.83$ & $35.30\pm5.64$ \\
SelectiveNet+$a$ & $8.28\pm1.58$ & $12.58\pm4.00$ & $67.98\pm2.06$ & $15.94\pm3.01$ & $40.54\pm7.52$ \\
Deep Gamblers+$a$ & $9.22\pm1.14$ & $12.48\pm4.06$ & $66.16\pm1.94$ & $14.98\pm3.54$ & $35.75\pm4.02$ \\
ConfidNet+$a$ & $6.46\pm0.55$ & $9.35\pm2.32$ & $67.44\pm1.23$ & $12.91\pm1.82$ & $32.87\pm4.43$ \\
MHP+$a$ & $11.46\pm1.19$ & $17.17\pm4.32$ & $66.86\pm2.03$ & $18.51\pm2.62$ & $45.06\pm4.93$ \\
CHASE-Fusion & \best{6.04\pm0.70} & \best{7.87\pm2.62} & $70.25\pm2.04$ & $16.15\pm1.94$ & $47.61\pm7.27$ \\
H-A & $6.11\pm0.49$ & $7.97\pm3.37$ & \best{71.61\pm2.51} & \best{19.45\pm3.98} & \best{55.55\pm8.90} \\
\bottomrule
\end{tabular*}

\vspace{0.55em}
\textit{Panel (c): Target coverage 80\%.}
\begin{tabular*}{\textwidth}{@{\extracolsep{\fill}}lccccc@{}}
\toprule
Method & $\Risk_{80}$(O) & $\Risk_{80}$(V) & $\Three_{80}$(O) & $\Three_{80}$(V) & $\Align_{80}$(O) \\
\midrule
MSP+$a$ & $4.36\pm0.89$ & $5.67\pm2.14$ & $66.13\pm1.34$ & $26.49\pm3.44$ & $34.46\pm5.16$ \\
MC Dropout+$a$ & $4.54\pm1.13$ & $6.82\pm3.87$ & $66.25\pm1.41$ & $27.21\pm2.96$ & $34.74\pm3.52$ \\
S-A & $4.35\pm1.10$ & $6.40\pm3.25$ & $66.43\pm1.35$ & $26.97\pm3.55$ & $35.06\pm4.57$ \\
SelectiveNet+$a$ & $5.61\pm1.43$ & $8.42\pm3.87$ & $66.73\pm2.64$ & $29.81\pm5.07$ & $37.45\pm4.46$ \\
Deep Gamblers+$a$ & $6.39\pm1.16$ & $8.99\pm3.23$ & $64.48\pm1.72$ & $27.41\pm3.96$ & $33.51\pm2.58$ \\
ConfidNet+$a$ & $4.41\pm0.89$ & $5.97\pm2.25$ & $65.94\pm1.71$ & $26.34\pm2.83$ & $34.14\pm5.05$ \\
MHP+$a$ & $8.50\pm1.70$ & $12.28\pm4.33$ & $67.74\pm2.53$ & $35.49\pm4.29$ & $44.84\pm3.54$ \\
CHASE-Fusion & \best{4.25\pm0.73} & $4.81\pm2.33$ & $71.18\pm2.01$ & $34.97\pm4.42$ & $48.12\pm5.91$ \\
H-A & $4.58\pm0.83$ & \best{2.94\pm1.70} & \best{76.18\pm2.30} & \best{46.06\pm3.87} & \best{63.40\pm6.62} \\
\bottomrule
\end{tabular*}
\end{table*}
\subsection{Meaningful ambiguity supervision and sensitivity}

The selector-target controls test whether ambiguity-aware abstention
depends on meaningful simulator ambiguity information (Supplementary
Section S1). All rows retain
the CHASE-Fusion representation and fusion and differ only in the target
used to supervise selection. Relative to the primary binary target
$a=\mathbb{1}[\alpha\geq0.75]$, error-only and shuffled-label
supervision reduce mean selective risk by 0.47--0.56~pp but decrease
three-way accuracy by 3.52--4.33~pp, abstain alignment by
10.99--13.10~pp, and ambiguity AUROC by 9.87--11.57~pp.
Random auxiliary targets therefore do not reproduce the
ambiguity-aware abstention gains.

The conclusion is stable under moderate changes to the binary ambiguity
cutoff. Across cutoffs from 0.65 to 0.80, the ranges are 0.22~pp for
risk, 0.29~pp for three-way accuracy, and 0.47~pp for alignment.
Performance weakens more noticeably at 0.85, while supervision with
continuous $\alpha$ produces an intermediate risk--alignment operating
point.

The margin ablation examines the class-separation margin $m$ in the
hypothesis-backbone objective (Supplementary Table~S2). Increasing $m$ requires the correct
hypothesis to outperform the incorrect hypothesis by a larger
predictive-loss margin. Moving from $m=0$ to the selected value
$m=1$ improves raw-margin no-abstain accuracy by 3.57~pp and reduces
risk by 3.31~pp. For H-A, $m=0.5$ increases three-way accuracy by
3.15~pp and alignment by 8.69~pp relative to $m=1$, but raises risk by
0.82~pp. Increasing the margin to $m=2$ lowers risk by a further
0.26~pp while reducing alignment by 15.81~pp. The selected value
$m=1$ therefore represents a balanced operating point rather than the
optimum of any single metric.

\begin{table*}[pos=t,width=\textwidth]
\centering
\caption{\textbf{Leave-one-regime-out results at the validation threshold nominally targeting 80\% coverage.} Achieved coverage and actual accepted-set risk are shown because transferred thresholds do not preserve nominal coverage. $\Risk_{80}$(O) is recomputed at a standardized 80\% ranking cutoff. Because three-way accuracy and abstain alignment also depend on achieved
coverage, their transferred-threshold values are descriptive and are not used
for formal cross-method inference. Standardized risk at 80\% coverage and the
complete risk--coverage curves provide the common-budget comparisons. Values are percentages, mean $\pm$ sample standard deviation over ten simulator seeds.}
\label{tab:ood_generalization}
\small
\renewcommand{\arraystretch}{1.04}
\setlength{\tabcolsep}{2pt}

\textit{Panel (a): Train intermittent; test local - prediction and risk.}
\begin{tabular*}{\textwidth}{@{\extracolsep{\fill}}lcccc@{}}
\toprule
Method & $\NA$(O) & Achieved cov. & Actual $\Risk$ & $\Risk_{80}$(O) \\
\midrule
Deep Ensemble & $78.46\pm3.59$ & $72.08\pm4.16$ & $12.55\pm2.19$ & $15.26\pm2.80$ \\
Raw margin & $\best{83.79\pm2.45}$ & $58.62\pm4.87$ & $4.85\pm2.28$ & $\best{10.31\pm2.48}$ \\
S-A & $78.12\pm3.17$ & $72.04\pm4.52$ & $11.66\pm2.15$ & $14.37\pm3.45$ \\
H-A & $\best{83.79\pm2.45}$ & $73.17\pm5.65$ & $9.04\pm2.68$ & $10.94\pm2.69$ \\
CHASE-Fusion & $79.79\pm3.11$ & $70.38\pm6.52$ & $10.53\pm4.22$ & $13.28\pm3.33$ \\
\bottomrule
\end{tabular*}

\vspace{0.4em}
\textit{Panel (b): Train intermittent; test local - ambiguity-aware utility.}
\begin{tabular*}{\textwidth}{@{\extracolsep{\fill}}lccc@{}}
\toprule
Method & $\Three$(O) & $\Align$(O) & Amb.-AUROC \\
\midrule
Deep Ensemble & $55.54\pm4.07$ & $25.76\pm5.46$ & $50.62\pm4.86$ \\
Raw margin & $56.08\pm4.11$ & $28.73\pm6.45$ & $57.64\pm5.56$ \\
S-A & $56.46\pm4.69$ & $26.97\pm6.83$ & $53.99\pm5.11$ \\
H-A & $\best{71.88\pm5.18}$ & $\best{55.85\pm10.17}$ & $\best{78.31\pm6.35}$ \\
CHASE-Fusion & $63.46\pm5.99$ & $41.49\pm14.13$ & $71.28\pm7.42$ \\
\bottomrule
\end{tabular*}

\vspace{0.6em}
\textit{Panel (c): Train local; test intermittent - prediction and risk.}
\begin{tabular*}{\textwidth}{@{\extracolsep{\fill}}lcccc@{}}
\toprule
Method & $\NA$(O) & Achieved cov. & Actual $\Risk$ & $\Risk_{80}$(O) \\
\midrule
Deep Ensemble & $77.88\pm2.96$ & $72.92\pm4.00$ & $15.02\pm2.18$ & $16.56\pm2.70$ \\
Raw margin & $\best{79.83\pm2.72}$ & $88.25\pm3.50$ & $16.64\pm3.32$ & $\best{13.33\pm2.80}$ \\
S-A & $78.38\pm2.82$ & $72.79\pm4.64$ & $13.56\pm2.94$ & $15.52\pm2.83$ \\
H-A & $\best{79.83\pm2.72}$ & $81.62\pm5.21$ & $14.23\pm3.15$ & $13.70\pm3.11$ \\
CHASE-Fusion & $79.42\pm3.99$ & $74.83\pm7.60$ & $12.95\pm3.75$ & $14.06\pm4.13$ \\
\bottomrule
\end{tabular*}

\vspace{0.4em}
\textit{Panel (d): Train local; test intermittent - ambiguity-aware utility.}
\begin{tabular*}{\textwidth}{@{\extracolsep{\fill}}lccc@{}}
\toprule
Method & $\Three$(O) & $\Align$(O) & Amb.-AUROC \\
\midrule
Deep Ensemble & $54.21\pm3.94$ & $22.89\pm8.59$ & $51.60\pm5.74$ \\
Raw margin & $64.12\pm4.61$ & $47.10\pm10.42$ & $70.81\pm3.49$ \\
S-A & $55.62\pm2.94$ & $25.50\pm10.76$ & $56.24\pm6.38$ \\
H-A & $\best{69.04\pm4.14}$ & $\best{58.50\pm11.42}$ & $\best{82.85\pm6.83}$ \\
CHASE-Fusion & $58.96\pm8.11$ & $32.84\pm15.28$ & $65.41\pm11.51$ \\
\bottomrule
\end{tabular*}

\parbox{\textwidth}{\footnotesize
\textit{Note:} Actual-risk values correspond to different achieved
coverages and are therefore not ranked or bolded. Boldface marks the
best mean only for no-abstain accuracy, standardized $\Risk_{80}$,
three-way accuracy, abstain alignment, and ambiguity AUROC. The results
show a tradeoff: the raw hypothesis margin provides the strongest
standardized risk ranking, whereas H-A provides the strongest
ambiguity-aware utility under regime shift.
}
\end{table*}

\subsection{Unseen-regime generalization}
Table~\ref{tab:ood_generalization} shows that the ambiguity-aware advantage persists when the test mechanism is absent from training. Because achieved coverage differs across transferred thresholds, the reported
three-way accuracy, abstain alignment, and actual accepted-set risk are
descriptive. We therefore do not use these transferred-threshold quantities
for formal cross-method inference. Common-budget risk comparisons use the
standardized \(R_{80}\), supplemented by complete risk--coverage curves. At the transferred 80\% threshold, H-A improves three-way accuracy over S-A by 15.42 pp and alignment by 28.88 pp for intermittent$\rightarrow$local, and by 13.42 and 33.00 pp for local$\rightarrow$intermittent. All four effects have the same direction on all ten simulator seeds and survive Holm correction. The raw hypothesis margin has the lowest standardized $\Risk_{80}$ in both directions, whereas H-A leads three-way accuracy, alignment, and ambiguity AUROC. Thus, direct margins remain effective for risk ranking, while the learned hypothesis selector better identifies which cases should be rejected as structurally ambiguous. At the transferred 90\% threshold, H-A retains three-way gains of 8.58 and 7.88 pp and alignment gains of 20.46 and 27.83 pp; its risk advantage over S-A is significant only for intermittent$\rightarrow$local.

The transferred thresholds do not preserve their nominal coverage: for
example, the raw margin attains 58.62\% coverage in
intermittent$\rightarrow$local but 88.25\% in
local$\rightarrow$intermittent at the nominal 80\% operating point.
This asymmetry arises because a threshold selected on one regime is calibrated
to that regime's accept-score distribution; when the scale or spread of scores
changes in the unseen regime, the same numerical threshold selects a different
fraction of cases. Consequently, the ``Actual R'' values in Table~\ref{tab:ood_generalization}
are evaluated at different achieved coverages and should not be compared as
though they shared a common operating budget. Fair risk comparisons therefore
rely on standardized $\Risk_{80}$ and complete risk--coverage curves.
These experiments demonstrate unseen-regime robustness within the simulator;
they do not establish general real-domain validity.

Complete risk--coverage curves for both leave-one-regime-out
directions are provided in Supplementary Figure~S2. These curves
compare the methods over common achieved-coverage values and complement
the transferred-threshold and standardized 80\%-coverage results in
Table~\ref{tab:ood_generalization}.

\subsection{Frozen-model real-GUV applicability}
\label{sec:real_guv_results}

\begin{figure}[pos=ht,width=\linewidth]
\centering
\includegraphics[width=\linewidth]{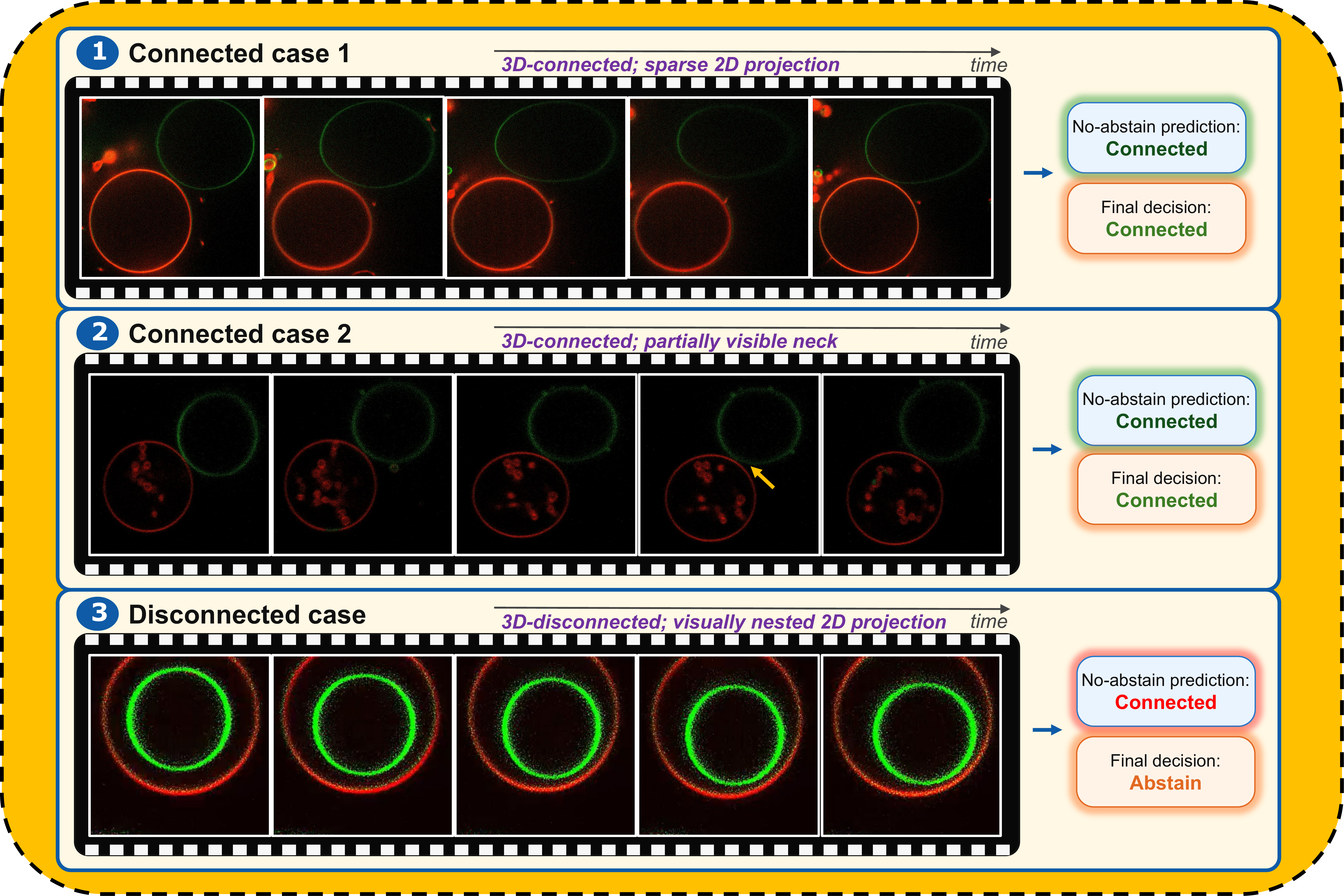}
\caption{\textbf{Representative real-GUV sparse-frame decisions from the frozen simulator-trained CHASE pipeline.} The two upper rows are physically connected examples and the lower row is the sole physically disconnected example, whose nested 2D projection is misleading. After label-blind geometric canonicalization, fold-consensus CHASE commits on the two connected examples and safely abstains on the disconnected example despite a forced binary prediction of Connected.}
\label{fig:real_GUV_main}
\end{figure}

CHASE-Fusion accepted 8 of 11 cases (72.7\% achieved coverage). All eight accepted consensus decisions agreed with the physical labels, giving an observed accepted-set risk of 0/8 (exact 95\% binomial interval: 0--36.9\%). The no-abstain branch was correct on 10 of 11 cases (90.9\%; exact 95\% interval: 58.7--99.8\%). Final decisions comprised eight Connected predictions and three Abstentions; no case received a final disconnected decision. The representative montage (Figure~\ref{fig:real_GUV_main}) shows two accepted connected examples and the abstained physically disconnected example. Notably, while the forced no-abstain prediction for the sole disconnected case was Connected due to deceptive 2D visual nesting, fold-consensus CHASE safely abstained. The frozen selector therefore committed on clear connected cases and abstained on the deceptive disconnected projection, avoiding an accepted physical-label error in this cohort. Full case-level outcomes and preprocessing details are reported in Supplementary Section~S5. Because the cohort contains ten connected cases and only one disconnected case, an always-connected rule also attains 90.9\% no-abstain accuracy. We therefore do not interpret no-abstain accuracy as evidence of real-domain binary discrimination; the relevant observation is that the sole disconnected case is rejected rather than accepted with an erroneous connected label. The pilot thus demonstrates selective behaviour on this development-inspected cohort, rather than reliable binary recognition of disconnected real cases.

% These findings are exploratory rather than confirmatory. Only one case is disconnected, sparse selected frames are interpolated to the temporal input length, and the engineered canonicalization is not an end-to-end learned representation. We therefore use the real-GUV result to demonstrate selective behaviour and observed accepted-set physical-label correctness on a small pilot, not superiority over the contextual baselines or general real-domain validity. Full case-level results and preprocessing details are reported in Supplementary Section S5.

\subsection{Computational cost}

The isolated CHASE-Fusion five-fold implementation was run on one NVIDIA
RTX A5000 GPU with a batch size of 64. Training required
$30.68\pm0.35$~s per fold, and scoring one 672-sequence test fold required
$1.46\pm0.05$~s ($2.17\pm0.08$~ms per sequence). Values are wall-clock
means $\pm$ sample standard deviations across five group-disjoint folds.

\section{Discussion}
\label{sec:discussion}

The experiments identify a representation effect that conventional uncertainty scores do not reproduce. Competing class-conditional predictors expose whether one temporal explanation consistently accounts for the observed sequence better than its alternative. This signal is already useful before ambiguity supervision is introduced, and matched-supervision comparisons show that the advantage persists when standard and hypothesis representations receive the same selector target. The contribution is therefore not access to an extra label alone, but a representation that makes structural insufficiency easier to rank. Statistical inference across simulator seeds quantifies reproducibility conditional on the chosen simulator family; it does not establish robustness to alternative data-generating mechanisms.

The results also reveal a deliberate risk--utility tradeoff. Error-only objectives can obtain slightly lower accepted-set risk by concentrating rejection on likely mistakes, yet they spend less of the abstention budget on cases that are genuinely unresolved. Ambiguity supervision moves the operating point toward higher three-way utility and alignment. H-A makes this shift more effectively than S-A, especially on the VH subset, while CHASE-Fusion provides the original fused alternative. Neither configuration uniformly dominates every risk operating point, so method choice should depend on whether the application values the lowest possible accepted-set error or more faithful identification of structurally ambiguous cases.
In GUV connectivity inference, this tradeoff corresponds to deciding whether
a visually unresolved two-dimensional projection should be routed for further
review rather than forced into a binary physical-connectivity label.

Independent datasets and unseen-regime transfer reduce two important threats to the original evidence. The replicated gains are not an artifact of five partitions of a single generated dataset, and they persist when the test ambiguity mechanism is absent from training. At the same time, transferred thresholds show that nominal coverage is not stable under shift. A deployed selector would therefore require coverage monitoring, post-shift recalibration, or an external risk-control layer.

% \textcolor{red}{check again.}The inferential benchmark is simulator-based and the ambiguity target is defined by simulator parameters. The real-GUV applicability cohort comprises 11 three-dimensionally validated sparse-frame cases and was inspected during development of the real-to-simulator preprocessing pathway.  A fine or out-of-plane connecting neck can be unresolved in an individual two-dimensional projection; prior GUV experiments likewise show that apparently separated lobes can remain membrane-connected through variable-diameter necks \cite{mathivet1996shape}. Moreover, osmotic stress can generate heterogeneous and time-dependent remodeling, including budding, fission, fusion, tubulation, and pearling \cite{oglecka2012osmotic}. Acquiring a balanced cohort of three-dimensionally validated connected and disconnected outcomes is therefore experimentally demanding. Accordingly, this cohort supports a frozen-transfer demonstration and case-level selective behaviour, but not population-level accuracy, class-balanced performance, or superiority claims. A prospectively collected, untouched cohort containing additional disconnected cases is required for confirmatory evaluation. The real-GUV transfer currently relies on a deterministic, label-blind geometric canonicalization to map fluorescence images to the six simulator-compatible descriptors. Because this preprocessing pathway was developed using the pilot cohort, prospective evaluation should freeze both frame selection and canonicalization before application to new three-dimensionally validated cases.
The inferential benchmark is simulator-based, and the ambiguity target is defined by simulator parameters. The real-GUV applicability cohort comprises 11 three-dimensionally validated sparse-frame cases and was inspected during development of the real-to-simulator preprocessing pathway. The small and imbalanced cohort reflects two sequential experimental constraints. First, only phase-separated GUVs that developed a constricted closed-neck morphology under hypertonic stress meet our requirements. In our experiments, such cases account for only about 50\% of the GUVs. In addition, the GUVs continuously drift due to the Marangoni flow induced by the hypertonic solution, which is a convective flow driven by concentration gradients. As a result, some GUVs occasionally move out of the field of view, making it difficult to track the morphology of individual GUVs throughout the entire deformation process.
Overall, after excluding GUVs that fail to form a closed neck or drift out of the field of view, only a limited number of videos remain suitable for analysis. Second, under the present protein-free osmotic-stress conditions, the GUVs commonly reached budded or narrow-necked states but only infrequently completed neck cleavage. Narrow-necked budded morphologies arise from the energetic balance among
membrane bending, domain-boundary line tension, and vesicle geometry
\cite{julicher1993domain}, whereas complete neck scission is a distinct
topological transition that requires overcoming a free-energy barrier
\cite{bottacchiari2022activation}. Therefore, in our experiments, the energy purely provided by osmotic imbalance, without proteins or ATPs assistance, is not enough to cleave the closed neck or to provide more disconnected cases. Accordingly, the cohort supports only a case-level frozen-applicability demonstration. Confirmatory evaluation requires a prospectively collected, untouched cohort containing additional three-dimensionally validated disconnected cases, with frame selection and geometric canonicalization fixed before model application.

Moreover, the current hypotheses are class-conditional predictive heads rather than fully interpretable physical models, and their candidate set is specified by the task: the two heads are semantically anchored to the known Connected and Disconnected labels. This is the same binary decision space used by the single-branch baselines, so the controlled comparisons test how a shared task definition is represented rather than giving CHASE access to a different prediction target. Nevertheless, the present results do not establish automatic discovery of the number or semantics of plausible dynamical explanations, nor do they show that two class-level hypotheses exhaust all modes of physical variation. A more general extension could learn an overcomplete or adaptive bank of latent temporal hypotheses and associate the resulting specialized modes with task labels after learning.
% The hypotheses are class-conditional predictive heads rather than fully
% interpretable physical models, and establishing formal conditions under which
% the hypothesis margin yields calibrated selective risk remains an important
% direction for future work. Finally, all models use pre-extracted six-dimensional descriptors; end-to-end representation learning remains untested. CHASE should therefore be viewed as a decision-support and triage mechanism, not autonomous physical ground truth.

\section{Conclusion}
\label{sec:conclusion}
CHASE turns selective prediction into an explicit comparison between competing temporal explanations. The raw likelihood margin establishes that the representation itself carries useful abstention information, while H-A shows that the same representation uses ambiguity supervision more effectively than a standard ensemble. Independent replication, label controls, sensitivity analysis, and unseen-regime transfer support a focused conclusion: CHASE improves the placement of abstentions---most clearly on very-high-ambiguity cases---at a competitive selective-risk boundary. The real-GUV applicability pilot further shows that the frozen selector can commit on clear connected examples while abstaining on a misleading disconnected projection, but the small, imbalanced development cohort and engineered sparse-frame canonicalization preclude a general real-domain claim. Independent held-out real cases, especially additional disconnected examples, and explicit post-shift coverage control remain necessary before deployment. Future work should relax the fixed hypothesis set by learning multiple latent dynamical modes---potentially with an adaptive number of modes---while retaining abstention as a decision about insufficient comparative evidence rather than treating it as an additional physical class.

\section*{Acknowledgments}
This research is supported by the Ministry of Education, Singapore, under its Research Centre of Excellence award to the Institute for Digital Molecular Analytics \& Science, NTU (IDMxS, grant: EDUNC-33-18-279-V12). The authors acknowledge Nadra Ashley and Klynen Kwok for assisting Y.G. with dataset curation.

\section*{Author Contributions}
K.J: Conceptualization, Methodology, Software, Formal analysis, Investigation, Visualization, Writing—original draft.\\
Y.G: Investigation, Data curation, Validation, Writing—review and editing.\\
A.N.P: Resources, Supervision, Writing—review and editing.\\
L.W: Conceptualization, Methodology, Supervision, Project administration, Funding acquisition, Writing—review and editing.

\section*{Declaration of competing interest}
The authors declare that they have no known competing financial interests or personal relationships that could have appeared to influence the work reported in this paper.

\section*{Data and code availability}
Code will be released publicly upon publication. Raw real-GUV videos will be shared subject to institutional data-sharing constraints. 

\section*{Declaration of generative AI and AI-assisted technologies in the manuscript preparation process}
During the preparation of this work, the authors used ChatGPT by OpenAI to assist with language refinement and document organization. The authors independently verified, revised, and approved all scientific content, analyses, references, figures, and conclusions and take full responsibility for the content of the published article.

\FloatBarrier
\bibliographystyle{cas-model2-names}
\bibliography{references}
\end{document}